%% file: speller.tex
\pdfoutput=1

\documentclass[11pt]{article}

\usepackage{emnlp2021}

\usepackage{times}
\usepackage{latexsym}
\usepackage{amsmath}
\usepackage{times}
\usepackage{cleveref}
\usepackage{tabularx}
\usepackage{graphicx}
\usepackage{booktabs}
\usepackage{float}
\usepackage{listings}
\usepackage{mathtools}
\usepackage{multirow}
\usepackage{amsfonts,amssymb}
\usepackage{xcolor}
\usepackage{tabularx}
\usepackage{multirow}
\usepackage{booktabs}
\usepackage{dsfont}
\usepackage{bm}
\usepackage{enumitem}
\usepackage{amsthm}
\usepackage{array}
\usepackage{anyfontsize}
\usepackage{bbm}
\usepackage{fixltx2e}
\usepackage{arydshln}

\usepackage[T1]{fontenc}

\usepackage[utf8]{inputenc}

\usepackage{microtype}
\usepackage{xcolor}

\title{Hierarchical Character Tagger for Short Text Spelling Error Correction}

\author{Mengyi Gao \thanks{\hspace{1mm} Equal Contribution} \\
  eBay Inc. \\
  \texttt{menggao@ebay.com} \\\And
  Canran Xu \footnotemark[1]\\
  eBay Inc. \\
  \texttt{canxu@ebay.com} \\\And 
  Peng Shi \footnotemark[1]\\
  eBay Inc. \\
  \texttt{pshi@ebay.com}}

\begin{document}
\maketitle

\input{src/abstract}
\input{src/intro}
\input{src/approach}

\input{src/experiments}
\input{src/related_works}
\input{src/conclusions}
\input{src/acknowledgement}

\bibliography{anthology, custom}
\bibliographystyle{acl_natbib}

\end{document}

%% file: src/abstract.tex
\begin{abstract}
State-of-the-art approaches to spelling error correction problem include Transformer-based Seq2Seq models, which require large training sets and suffer from slow inference time; and sequence labeling models based on Transformer encoders like BERT, which involve token-level label space and therefore a large pre-defined vocabulary dictionary. In this paper we present a Hierarchical Character Tagger model, or HCTagger, for short text spelling error correction. We use a pre-trained language model at the character level as a text encoder, and then predict character-level edits to transform the original text into its error-free form with a much smaller label space. For decoding, we propose a hierarchical multi-task approach to alleviate the issue of long-tail label distribution without introducing extra model parameters. Experiments on two public misspelling correction datasets demonstrate that HCTagger is an accurate and much faster approach than many existing models. 
\end{abstract}

%% file: src/intro.tex
\section{Introduction}

A spelling corrector is an important and universal tool for a wide range of text-related applications, such as search engines, machine translation, optical character recognition, medical records, text processors and essay scoring. Although spelling error correction is a long-studied problem, it remains a challenging task because words can be misspelled in a variety of forms, including in-word errors, cross-word errors, non-word errors and real-word errors, depending on the subtle contextual information.

\begin{figure}[ht]
    \centering
    \includegraphics[width=0.45\textwidth]{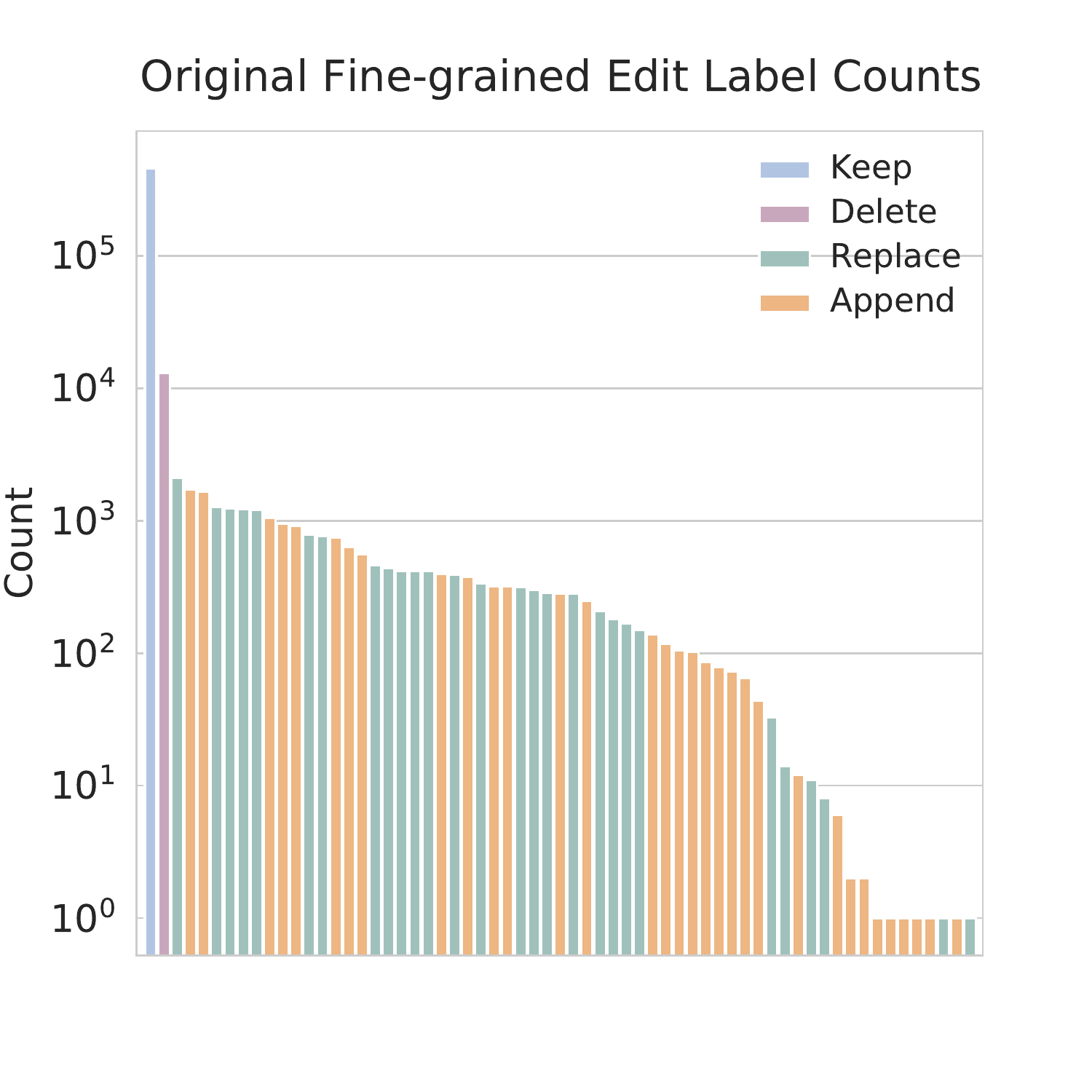}
    \includegraphics[width=0.45\textwidth]{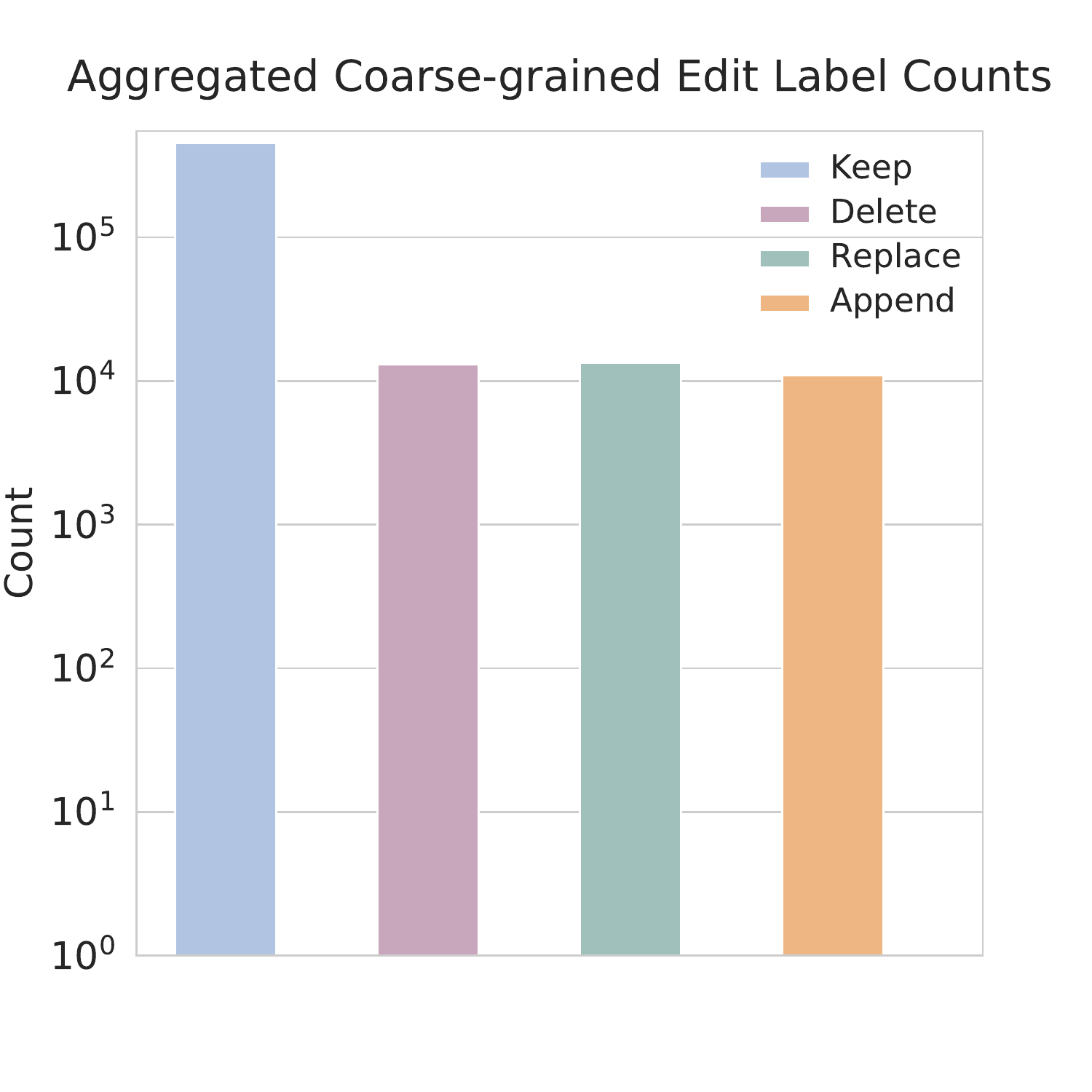}
    \caption{Original and aggregated edit label counts. The upper plot shows original fine-grained edit label counts, which are heavily skewed. The lower plot of aggregated coarse-grained edit label counts has much less skewness.}
    \label{fig:edit_label_cnt}
\end{figure}

In this paper, we focus on solving the spelling correction problem in user-generated short text, such as queries in search engines or tweets on social media, which has three unique properties compared to long essays. First, search queries or tweets are often short and lack context. Second, most short text contains pure spelling errors and almost no grammatical errors. Third, instant spell checkers used in search engines or social medias have strict latency requirements.   


In general, popular approaches to spelling correction make use of parallel corpora in which the source sentence contains spelling errors and the target sentence is error-free. Recently, the Transformer-based sequence-to-sequence (Seq2Seq) model \cite{transformer} has gradually proven to be effective on spelling correction problems. Unlike neural machine translation, spelling errors tend to occur locally for a few characters while the rest of the text is correct. To cope with this situation, \citeauthor{GEC_copy} propose a scheme to incorporate a copy mechanism in Seq2Seq. The success of this type of Seq2Seq model depends on large scale annotated datasets, which are often generated by constructing text noise from clean text in previous studies. Moreover, this approach suffers from slower inference time and lack of interpretability.

Another  class  of  approaches  is  based  on  sequence labeling. Instead of generating the output sequence in an autoregressive fashion, PIE \cite{pie} and GECToR \cite{GECToR} predicts token-level edit operations in one of \{\textit{Keep}, \textit{Delete}, \textit{Replace}, \textit{Append}\} by leveraging pre-trained Transformer encoders, such as BERT \cite{bert}. Such models can generate the outputs for all tokens in parallel, and therefore significantly reduces the latency of sequential decoding as in Seq2Seq models while achieving comparable accuracy. However, the approaches in both papers predict edit operations at token level. It can be expected that the \textit{Replace} and \textit{Append} operations are associated with
a huge pre-defined vocabulary dictionary. For real-life usages it is infeasible to enumerate all correctly spelled words in the label space.

\begin{figure*}[ht]
    \centering
    \includegraphics[width=0.99\textwidth]{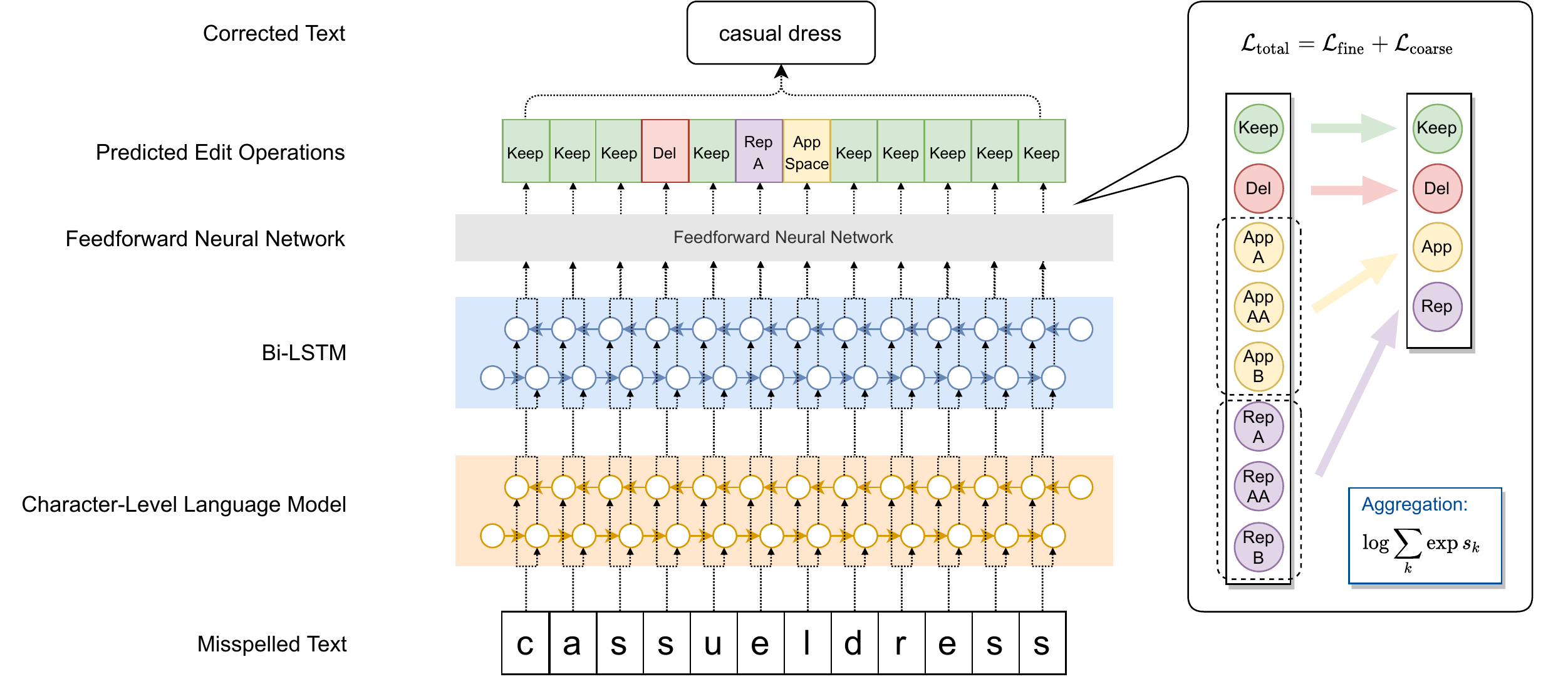}
    \caption{Overall architecture of the model. The text encoder is a character-level language model, followed by a bi-directional LSTM. The edit operations predicted by the feedforward neural network are used to formulate the corrected text. During training, the output of the feedforward neural network is used to construct the hierarchical loss function with two explicit terms.}
    \label{fig:model_arch}
\end{figure*}

To address the aforementioned shortcomings, considering the unique properties of short text misspelling correction, in this paper, we propose a new model called Hierarchical Character Tagger, or HCTagger for short, which uses a pre-trained language model at the character level as a text encoder, and then predicts character-level edits. It is motivated by the straightforward observation that spelling errors usually occur at character level.  For the misspelling-correction pair of \textit{shies} $\rightarrow$ \textit{shoes}, its character-level edit labels would be [\textit{s: Keep}, \textit{h: Keep}, \textit{i: Replace with \textbf{o}}, \textit{e: Keep}, \textit{s: Keep}], which is represented in a much smaller label space compared to [\textit{shies: Replace with \textbf{shoes}}] at token level. While most spelling errors occur within 1-edit distance for each token, for broader coverage we also include character sequence edit operations like \textit{Replace with \textbf{oa}}.

Furthermore, the distribution of edit labels is long-tailed. As shown in \Cref{fig:edit_label_cnt}, \textit{Keep} and \textit{Delete} are more frequent than labels of \textit{Replace/Append with certain character(s)}. If these labels are treated as equivalent, the overall accuracy of the model will be constrained by the unevenness of the label distribution. Therefore, for decoding, we aggregate original fine-grained edit labels into four coarse-grained labels [\textit{Keep}, \textit{Delete}, \textit{Append}, \textit{Replace}]. We propose a hierarchical multi-task approach to learn both the fine-grained and coarse-grained edit labels at the same time without introducing any extra model parameters.

Through extensive experiments on two public datasets, we demonstrate that our proposed HCTagger effectively improves the performance and latency of short text spelling correction.

%% file: src/approach.tex
\section{Approach}
We describe our model HCTagger in this section.

\subsection{Problem Formulation}
Without lack of universality of language types, for an input text sequence with spelling error, $S = [c_1, \dots, c_n]$,  our goal is to get the correct spelling of the corresponding text, denoted as $T = [d_1, \dots, d_m]$, where $c_i$ and $d_i$ are character level input and output. Note that the sequence lengths $n$ and $m$ are not necessarily equal. 

To map the source sequence $S$ to target $T$, a corresponding edit operation sequence $O = [o_1, \dots, o_n]$ is applied. Note that $O$ has the same sequence length as $S$. Each edit operation, $o_i$, falls into one of the following four categories:
\begin{description}[leftmargin=0cm]
\item[Keep] The current character remains unchanged. This means that the current character is not misspelled.
\item[Delete] The current character is deleted.
\item[Append] Append a sequence of characters of length greater than or equal to one after the current character. Each distinct appended sequence is treated as an independent tag type.
\item[Replace] Replaces the current character with a number of characters of length greater than or equal to 1. Similar to \textit{Append}, each distinct appended sequence is treated as an independent tag type. 
\end{description}

Note that there could be more than one possible edit operation sequences to transform from source $S$ to target $T$. We use Python function \textit{SequenceMatcher} in module \textit{difflib} to do obtain the unique edit operation label sequence $O$. The idea of \textit{SequenceMatcher} is to find the longest contiguous matching subsequence. This does not necessarily yield minimal edit sequences, but does tend to yield matches that "look right" to humans. For more information, refer to the doc \footnote{\url{https://docs.python.org/3/library/difflib.html}}.

Thus, we eventually transform spelling correction into a sequential labeling problem, i.e., for a given input $S = [c_1, \dots, c_n]$, predict the edit operation $o_i$ for each character $c_i$. As a concrete example, to map a misspelled input text \textit{cassueldress} to its correction \textit{casual dress}, 3 edit operations are required, namely (1) deleting the 4th character \textit{s}, (2) replacing the 6th character \textit{e} with \textit{a}, and (3) appending a \textit{space} after the 7th character \textit{l}, while keeping other characters unchanged.



\subsection{Model}
Our proposed model, HCTagger, consists of two components. First, we encode the text by pre-training a character-level language model. Second, the representation obtained by the language model is encoded by a bi-directional LSTM, which is then fed to a decoder. This decoder is hierarchical: it decodes simultaneously four coarse-grained labels [\textit{Keep}, \textit{Delete}, \textit{Append}, \textit{Replace}] and all the fine-grained tags (such as \textit{Append with \textbf{a}} or \textit{Replace with \textbf{eo}}), which could be of potentially up to thousands types. The architecture of the model is shown in \Cref{fig:model_arch}.

\begin{table*}[ht]
\centering
\small
\begin{tabular}{cccc}
\toprule
\# Iteration  & Short Text & \# Token-level Errors & \# Character-level Errors \\
\midrule
     Original & fashi{\color{red}\textbf{e}}n industr{\color{red}\textbf{ie}} & 2 & 3\\
     1 & fashi{\color{blue}\textbf{o}}n industr{\color{blue}\textbf{y}}  & 0 & 0\\
\bottomrule
\end{tabular}\caption{An illustrative example of iterative inference.}\label{tab:iteration}
\end{table*}

\vskip 0.1in
\noindent\textbf{Character-level Language Model} The character-level language model we use is the pretrained Flair \cite{flair}, which has been widely shown to be effective for word-level sequence labeling tasks. Specifically, Flair consists of a character-level embedding layer and a (possibly bi-directional) LSTM layer. The model predicts the next character by the preceding or succeeding character inputs. The authors argue that it can capture semantic differences in morphological similarities, as well as contextual information for polysemous words. Moreover, character-level models better handle rare and misspelled words as well as model subword structures such as prefixes and endings.

Though pre-training a character-level language model, in the original paper Flair focuses on word-level sequence labeling task (e.g., NER). Specifically, to obtain word-level embeddings from character-level language model, Flair uses the output hidden state after the last character in each word as the representation of the whole word. However, in our scenario, we use the embedding of current character to predict its own edit operation, regardless of which word it belongs to, even if it is a space or punctuation.

In addition, when using Flair as the text encoder, we found that fine-tuning Flair's language model parameters along with the sequence labeling task generally perform better than without fine-tuning. Therefore, fine-tuning Flair is our default setting whenever possible.

\vskip 0.1in
\noindent\textbf{Hierarchical Multi-Task}
In a training set of finite size, the original fine-grained edit labels (a certain character being appended or replaced with some characters) form a long-tail distribution, as shown in \Cref{fig:edit_label_cnt}. This makes some relatively rare spelling errors more difficult to be corrected.


For decoding, we feed the hidden states of the bidirectional LSTM into a layer of feedforward neural network whose output dimension is the size of label types. For character $c_i$, the probability of original fine-grained label type $k$ is $P(k | c_i)$.
To alleviate the issue of long-tail distribution for $k$, we propose to aggregate the probabilities for four coarse-grained edit labels, denoted as $P(v|c_i)$, with $v \in {\rm  \{\textit{Keep}, \textit{Delete}, \textit{Replace}, \textit{Append}\}}$, which are presumably more balanced than the fine-grained labels. To achieve this, we use the rule of sum of probability: as all possible fine-grained \textit{Append} (\textit{Replace}) operations are mutually exclusive, the sum of their probabilities should equal the coarse-grained probability of \textit{Append} (\textit{Replace}). Formally, 
\begin{equation}\label{eq:sum_prob}
    P(A(R)| c_i) = \sum_{k\in A(R)_{\subset}} P(k | c_i),
\end{equation}
where $A_{\subset}$ and $R_{\subset}$ are the subsets consisting of fine-grained \textit{Append} and \textit{Replace} operations, correspondingly. 

Denote the logits for original fine-grained tag type $k$ as $f_k$, and logits for aggregated coarse-grained tag type $v$ as 
$l_v$. Then the probability of label type $k$ is $P(k | c_i) = {\rm softmax} (f_k)$. Similarly we have $P(v | c_i) = {\rm softmax} (l_v)$.
Therefore, \Cref{eq:sum_prob} can be derived as:
\begin{equation}\label{eq:sum_prob_sm}
    \frac{\exp{(l_{A(R)})}}{\sum_{m\in\{K, D, A, R\}} \exp{(l_m)}} = \sum_{k\in A(R)_{\subset}} \frac{\exp{(f_k)}}{\sum_j \exp{(f_j)}},
\end{equation}
where $K, D, A$ and $R$ are the short forms for \textit{Keep}, \textit{Delete}, \textit{Append} and \textit{Replace}, accordingly.

\begin{table*}[ht]
\small
\centering
\begin{tabular}{lccccc}
\toprule     
Dataset & \# Train & \# Dev & \# Test & \% Error Rate & \# Label Types \\
\midrule
Twitter &  31,172 & 4,000 & 4,000 & 100 & 66 \\

Webis & 44,772 & 5,000 & 5,000 & 17 & 112 \\

\bottomrule
\end{tabular}
\caption{Basic statistics of the datasets.}
\label{tab:dataset_stats}
\end{table*}

As a result, we obtain the coarse-grained logits $l_v$ by solving \Cref{eq:sum_prob_sm}:
\begin{equation}\label{eq:coarse}
l_v =\left\{
\begin{array}{ll}
f_k & {k = \textit{Keep}}\\
f_k & {k = \textit{Delete}}\\
\log\sum_{k\in A_{\subset}} \exp{(f_k)} & {k \in A_{\subset}} \\
\log\sum_{k\in R_{\subset}} \exp{(f_k)} & {k \in R_{\subset}} \\
\end{array} \right.
\end{equation}

Finally, HCTagger is trained by using the following multi-task loss associated with predicting edit $o_i$ at each character $c_i$:
\begin{align*}
    \mathcal{L} &= \mathcal{L}_{\rm fine} + \mathcal{L}_{\rm coarse} \\
    & =-\sum_i \log P(f_k^{(i)} | c_i) - \sum_i \log P(l_v^{(i)} | c_i).
\end{align*}

Notice that, in contrast to traditional multi-task learning, with the relation between $l_v$ and $f_k$ in \Cref{eq:coarse}, the coarse-grained loss function we introduce as an auxiliary task does not contain extra model parameters. The advantage of this design is that both fine-grained and coarse-grained loss functions can reach the optimum at the same time without additional efforts to tune the parameters to balance the two terms.

\vskip 0.1in
\noindent\textbf{Inference} Some previous studies \cite{pie, GECToR} on Grammar Error Correction (GEC) have shown that a well-established approach for inference is iterative: use the modified result obtained by the model in the current round as input for next round's prediction, and repeat the process several times. These studies find that due to the dependency among grammatical errors (tense, pronoun, subject-verb, preposition, plurals), the performance of model predictions can be steadily improved by multiple iterations. However, iterative inference is confronted with a trade-off between speed and accuracy.

As spelling errors impose a strong notion of locality and have weaker dependency on each other than grammatical errors, the iterative correction process is less necessary. For example, in the example shown in \Cref{tab:iteration}, the 2 token-level typos, \textit{fashien} and \textit{industrie}, 
 are independent of each other. To this end, the two errors can be corrected simultaneously through a single run in our model. Indeed, from our experiments, we noticed that having more than one round of inference iteration only marginally improves the accuracy in our task. 
Therefore, we report the results for HCTagger with only \textit{one} inference iteration in all experiments.

%% file: src/experiments.tex
\section{Experiments}
In this section, we describe the experiments performed on two public datasets for HCTagger. Meanwhile, we compare it with several state-of-the-art baselines.

\subsection{Datasets}
We conduct experiments on the following two short text datasets:
\begin{description}[leftmargin=0cm]
\item[Twitter Dataset] is proposed in \citet{twitter_dataset}, which includes 39,172 samples in their spell-error form and error-free form. We adopt the same train / dev / test split as \citet{twitter_char_label} and \citet{pie}.
\item[Webis Dataset] is introduced in \citet{webis}, which consists of 54,772 queries from AOL search logs. In contrast to the Twitter Dataset, the error rate of this dataset is only $\sim$17\%. Since the original dataset does not provide the train / dev / test split, we randomly sample 5000 queries as the dev and test sets, respectively, and use the remaining data as the training set.
\end{description}

The basic statistics of these two datasets and the corresponding number of label types (calculated from training data) are listed in the \Cref{tab:dataset_stats}.

\subsection{Baselines and Implementation Details}
The following baseline models are used for the comparison experiments:
\begin{description}[leftmargin=0cm]
\item[Aspell] \cite{aspell} works at word level. It uses a combination of metaphone phonetic algorithm, Ispell's near miss strategy and a weighted edit distance metric to score candidate words.
\item[Seq2Seq-LSTM] is the standard LSTM-based Seq2Seq architecture.
\item[Seq2Seq-Transformer] \cite{transformer} is the self-attention based Seq2Seq model.
\item[Local Sequence Transduction] \cite{twitter_char_label} treats spelling correction as a character-level local sequence transduction task by first predicting insertion slots, followed by a sequence labeling task for output tokens or a special operation \textit{Delete}.
\item[BERT-PIE] \cite{pie} or Parallel Iterative Edit model, is a sequence labeling method which uses BERT as its text encoder.
\item[BERT-Neuspell] \cite{neuspell} is provided by the Neuspell toolkit. It regards spelling correction as a token-level sequence labeling task, where the output for each token is its error-free form. We finetune the BERT model on the Webis dataset.
\end{description}

All the models are implemented in PyTorch \cite{pytorch}, and trained with a single Tesla V100 GPU. For HCTagger, we use the English Flair embeddings pretrained on the 1-billion word corpus \cite{flair_corpus}, which are publicly available\footnote{\url{https://github.com/flairNLP/flair}}. We tune the number of LSTM hidden states $\in\{512, 1024\}$, training batch size $\in\{8, 16, 32\}$, learning rate $\in\{1e^{-2}, 1e^{-3}\}$, and optimizer type $\in$\{Adam \cite{adam}, LAMB \cite{lamb}\}. In addition, both the encoder and decoder of Transformer has two self-attention layers.

\begin{table*}[ht]
\centering
\small
\begin{tabular}{lcc}
\toprule     
Model & Twitter Dataset Accuracy & Webis Dataset Accuracy\\
\midrule
Aspell & 30.1 $^{\dagger}$ & 65.8 \\
Seq2Seq (LSTM) & 52.2 $^{*}$ & 83.5\\
Seq2Seq (Transformer) & \textbf{67.6} $^{*}$ & 83.7 \\
\citet{twitter_char_label}  & 64.6 $^{\dagger}$ & - \\
BERT-PIE \cite{pie}  & 67.0 $^{*}$ & - \\
BERT-Neuspell \cite{neuspell} & - & 84.0 \\ 
\midrule
HCTagger & 67.2 \hspace{1.5mm} & \textbf{86.8} \\

\bottomrule
\end{tabular}
\caption{Performance on Twitter and Webis dataset. Results with $\dagger$ are from \citet{twitter_char_label}; results with $*$ are from \citet{pie}. }
\label{tab:perf}
\end{table*}



\begin{table}[ht]
\centering
\small
\begin{tabular}{lc}
\toprule     
Model &  Words per Second \\
\midrule
Seq2Seq (Transformer) & 36.62 \\
BERT-PIE \cite{pie} & 80.43 \\
\midrule
HCTagger & \textbf{251.20} \\
\bottomrule
\end{tabular}
\caption{Inference speed on Twitter dataset.}
\label{tab:twitter_speed}
\end{table}

\begin{table}[ht]
\centering
\small
\begin{tabular}{lc}
\toprule     
Model & Query per Second \\
\midrule
Seq2Seq (LSTM) & 83.33 \\
Seq2Seq (Transformer) & 40.00 \\
BERT-Neuspell \cite{neuspell} & 62.50 \\
\midrule
HCTagger & \textbf{250.00} \\
\bottomrule
\end{tabular}
\caption{Inference speed on Webis dataset.}
\label{tab:webis_speed}
\end{table}

\subsection{Results}

For the Twitter dataset, to align with previous publications, we report the accuracy in the test set to compare the performances among all models. As shown in \Cref{tab:perf}, HCTagger improves accuracy over all the models except Transformer. In particular, it is important to note that although the pretrained language model (Flair) we use is lightweight compared to BERT, our model still outperforms BERT-PIE. 

\Cref{tab:perf} also reports the performance on the Webis dataset. Our HCTagger exceeds all other models. Transformer model doesn't perform well on this dataset, probably because the number of misspelled queries is small (17\%) and is not enough to train Transformer well. In contrast, our model makes more effective use of small training set.

Meanwhile, we also compare the inference speed of the most accurate models, as shown in \Cref{tab:twitter_speed} and \Cref{tab:webis_speed}. Indeed, the inference speed of HCTagger is much faster than Seq2Seq (LSTM, Transformer), BERT-PIE \cite{pie}, and BERT-Neuspell \cite{neuspell}.  

\subsection{Ablation Study}
To understand the importance of each part of the model, we conduct an ablation study on the Twitter dataset, and report the accuracy in \Cref{tab:ablation}.

We first take away the pre-trained language model. At this point, the character-level embedding is randomly initialized and the rest of the model is left unchanged. The accuracy decreases by 1.9\%.

Subsequently, we preserve the language model but remove the coarse-grained loss term of the Hierarchical Multi-Task. In this case, the accuracy decreases by 0.7\%. 

\begin{table}[ht]
\centering
\small
\begin{tabular}{lc}
\toprule     
 & Accuracy \\
\midrule
Full Model & 67.2 \\

 - w/o Pretrained LM & 65.3 \\
 - w/o Hierarchical Multi-Task & 66.5 \\
\bottomrule
\end{tabular}
\caption{Ablation study on Twitter dataset.}
\label{tab:ablation}
\end{table}

%% file: src/related_works.tex
\section{Related Works}
\citeauthor{ebay_speller} use character-based statistical machine translation to correct user queries in the e-commerce domain. They extract training data from query refinement logs, and evaluate the results on an internal dataset.   

Grammar Error Correction (GEC) is an extensively researched NLP task.
This task contains grammar errors, including spelling, punctuation, grammatical, and word choice errors. PIE \cite{pie} and GECToR \cite{GECToR} are the state-of-the-art models that predict token-level edit operations \{\textit{Keep}, \textit{Delete}, \textit{Replace}, \textit{Insert}\} by leveraging pre-trained Transformer encoders like BERT. However, their models are not specifically designed for correcting spelling errors, which most often occur at character level. They rely on a small ($\sim$1k) pre-defined token-level \textit{Replace} and \textit{Insert} dictionary. Including all correctly-spelled tokens in the dictionary will make the label space too large.

Transformer based Seq2Seq models \cite{gec_corpora_generation_seq2seq, GEC_copy} prove to be successful on grammar error corrections, but heavily depends on synthetically generated error datasets. Character based Seq2Seq models \cite{gec_char_seq2seq} are also explored. Such model architectures involve a separate autoregressive decoder and attention module, which makes the inference time much slower. In particular for spelling error correction task, where the misspelling and correction only differ by one or more characters, Seq2Seq models seem too heavy.

Neuspell \cite{neuspell} is a spelling correction toolkit, which implements a wide range of models like SC-Elmo-LSTM and BERT. They regard spelling correction as a token-level sequence labeling task, where the output for each token is its error-free form.
For each word in the input text sequence, models are trained to output a probability distribution over a finite vocabulary.
Besides the aforementioned excessive label space problem at token-level, another shortcoming of this toolkit is that it assumes the misspelled and correction sentences have exactly the same number of tokens. Therefore, cross-word errors such as \textit{power point} $\rightarrow$ \textit{powerpoint} or \textit{babydoll} $\rightarrow$ \textit{baby doll} cannot be handled properly.

\citeauthor{twitter_char_label} treat spelling correction as a character-level local sequence transduction task by first predicting insertion slots in the input using learned insertion patterns, and then using a sequence labeling task to output tokens or a special token \textit{Delete}. They maintain a dictionary to keep track of the insertion context. For example, letter \textit{a} is inserted frequently after letter \textit{s}. While our pre-trained language model layer implicitly encodes such insertion context without the need of keeping a dictionary.

%% file: src/conclusions.tex
\section{Conclusions}
We presented the Hierarchical Character Tagger to correct user-generated short text misspellings. HCTagger predicts character-level edits, which has smaller label space than token-level edits. Pre-trained character-level language model embedding that we use is lightweight and much faster than BERT-like text encoders in many other state-of-the-art models, while achieving similar or even higher accuracy for short text spelling error correction task. Moreover, our novel Hierarchical Multi-Task decoding approach can be extended to any scenario that contains a hierarchical long-tail distributed label space. 

%% file: src/acknowledgement.tex
\section*{Acknowledgements}
We would like to thank Zhe Wu, Julie Netzloff, Xiaoyuan Wu, Hua Yang, Vivian Tian and Scott Gaffney for their support. 